# A Framework for Evaluating the Impact of Food Security Scenarios

Rachid Belmeskine
Convergaince Corporation
One Broadway, Kendall Square
Cambridge, MA 02142
rachid@convergaince.com

Abed Benaichouche
Convergaince Corporation
One Broadway, Kendall Square
Cambridge, MA 02142
abed@convergaince.com

*Abstract—* This study proposes an approach for predicting the impacts of scenarios on food security and demonstrates its application in a case study. The approach involves two main steps: (1) scenario definition, in which the end user specifies the assumptions and impacts of the scenario using a scenario template, and (2) scenario evaluation, in which a Vector Autoregression (VAR) model is used in combination with Monte Carlo simulation to generate predictions for the impacts of the scenario based on the defined assumptions and impacts. The case study is based on a proprietary time series food security database created using data from the Food and Agriculture Organization of the United Nations (FAOSTAT), the World Bank, and the United States Department of Agriculture (USDA). The database contains a wide range of data on various indicators of food security, such as production, trade, consumption, prices, availability, access, and nutritional value. The results show that the proposed approach can be used to predict the potential impacts of scenarios on food security and that the proprietary time series food security database can be used to support this approach. The study provides specific insights on how this approach can inform decision-making processes related to food security such as food prices and availability in the case study region.

## 1. INTRODUCTION

Food security is a complex issue that is influenced by a range of factors, including production, trade, and consumption patterns [1, 2, 3]. Ensuring food security is a global challenge [4], as it requires addressing the needs of a growing and increasingly diverse population while also ensuring the sustainability of the food system. The inability to predict the impacts of the various factors that influence food security is a major challenge, as it requires modeling the complex interactions between these factors and their dynamic nature over time. In this study, we aim to address this challenge by proposing an approach for predicting the impacts of scenarios on food security and demonstrate its application in a case study. The case study region is the United Arab Emirates (UAE) which was selected due to the high dependence of the UAE on food imports, and the need to enhance food security in this country.

Traditional approaches for predicting the impacts of food security factors, such as statistical models and expert judgment, have limitations in terms of accuracy, flexibility, and scalability. Machine Learning (ML) and Monte Carlo simulation-based approaches can overcome these limitations and enable the prediction of the impacts of food security scenarios with greater accuracy and flexibility.

The proposed approach involves two main steps: (1) scenario definition, in which the end user specifies the assumptions and impacts of the scenario using a scenario template, and (2) scenario evaluation, in which a Vector Autoregression (VAR) model [5] is used in combination with Monte Carlo simulation [6] to generate predictions for the impacts of the scenario based on the defined assumptions and impacts.

The structure of this paper is as follows: In the next section, we will provide a state of the art on existing approaches for predicting the impacts of scenarios on food security. Then, we will present the methods used in our proposed approach, including the creation of our proprietary time series food security database, the scenario definition and evaluation process, and the use of the Vector Autoregression (VAR) model in combination with Monte Carlo simulation. We will then present and discuss the results of our case study in the UAE. Finally, we will provide a conclusion on the potential of our proposed approach to enhance the decision-making processes related to food security.

## 2. STATE OF THE ART

Predicting the impacts of scenarios on food security has been a topic of interest for researchers and decision-makers [7, 8, 9]. Traditional approaches for predicting the impacts of food security factors include statistical models, expert judgment, and simulation models. Statistical models, such as linear regression and time series analysis, are based on the assumption of a linear relationship between the factors and the impacts [10, 11]. Expert judgment involves using the knowledge and experience of experts to predict the impacts of the factors [12, 13, 14]. Simulation models, such as system dynamics and agent-based models, are based on the representation of the relationships between the factors and the impacts using mathematical equations [15, 16, 17]. These approaches have several limitations in terms of accuracy, flexibility, and scalability. Statistical models are limited in their ability to capture the non-linear relationships between the factors and the impacts [18]. Expert judgment is subjective and depends on the expertise and experience of the experts and is prone to biases and errors [12]. Simulation models are complex and require the specification of the relationships between the factors and the impacts, which can be difficult and time-consuming [19]. In addition, these approaches are limited in their ability to handle the uncertainty and variability of the factors, and to generalize the predictions to new scenarios.

In recent years, Machine Learning and Monte Carlo simulation-based approaches have gained popularity for predicting the impacts of scenarios on food security due to their ability to overcome the limitations of traditional methods such as statistical models, expert judgment, and simulation models. These methods can capture the complex relationships between the factors and the impacts and handle the uncertainty and variability of the factors, enabling the prediction of the impacts of food security scenarios with greater accuracy and flexibility.

In [20], the study assesses the geographical pattern of association between food security and socioeconomic factors in urban areas in Iran using a household consumption-expenditure survey from 2010 to 2018 and a logistic regression model. A novel methodology combining machine learning and Monte Carlo simulations has been proposed in a study [21] that utilized EWF Nexus thinking to influence decision-making within the food sector and consider uncertainty related to energy prices derived from natural gas. This approach was used to assess technology alternatives for domestic food production in the State of Qatar based on economic and environmental performance. The Agrimonde-Terra foresight study [22] also utilized a scenario building method combining morphological analysis and quantitative simulations with a biomass balance model to explore the possible future of the global food system and consider a range of alternative diets with differing nutritional and health issues and urbanization and rural transformation processes. The SSPs [23] provide a consistent framework for studying the potential consequences of alternative global development pathways and land use patterns, but do not fully consider the potential for dietary changes and the resulting impacts on land use.

Conclusion: The traditional approaches for predicting the impacts of scenarios on food security have limitations in terms of accuracy, flexibility, and scalability. More recently, machine learning and Monte Carlo simulation-based approaches have gained popularity due to their ability to capture complex relationships and handle uncertainty and variability. However, these approaches are specific to the scope of previous studies and do not provide a generic solution for evaluating the impact of scenarios. The proposed approach in the current study aims to address these limitations and provide a more comprehensive approach for evaluating the impact of scenarios on food security.

## 3. METHODS

The methods section of this study proposes an approach for predicting the impacts of scenarios on food security. The approach consists of two main steps: (1) scenario definition and (2) scenario evaluation.

### 3.1. Time Series Food Security Database

In this study, a proprietary time series food security database was developed to support the prediction of the impacts of scenarios on food security. The database was created using data from multiple sources, including the Food and Agriculture Organization of the United Nations (FAOSTAT), the World Bank, and the United States Department of Agriculture (USDA). The database includes time series data on a wide range of indicators related to food security, including production, trade, consumption, prices, availability, access, and nutritional value. To ensure the quality and consistency of the data, a comprehensive data cleaning and preprocessing process was conducted. This process included steps such as removal of missing or duplicate data, handling of outliers, data transformation and terminology matching. This resulted in a high-quality and consistent database that can be used to support the prediction of the impacts of scenarios on food security.

The database includes data from various regions and countries, and covers a wide range of food items, such as cereals, fruits, vegetables, livestock, and fish. The data are available at various frequencies, such as annual, monthly, and daily. The database also includes various economic, social, and environmental indicators, such as GDP, population, poverty, and climate.

The proprietary time series food security database was created using advanced data management techniques such as data warehousing, data mining, and data integration. The data was organized in a structured format, linking each data point to specific attributes, including:

- Metric, which quantifies the data being measured, such as production, price, yields, etc.
- Item, which represents what the data is about, such as livestock (cow, camel, goat, etc.), fruits (apple, orange, etc.), vegetables, etc.
- Region, which identifies where the data originates, such as the country, region, city, farm. In some cases, a secondary region is also considered, for example, in the case of trade where we need to identify both the source and destination regions.
- Frequency, which represents the time period covered, such as annually, monthly, weekly, or daily

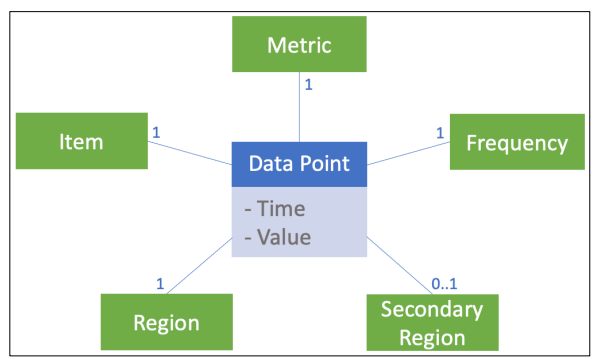

Figure 1. Data Model of the Food Security Database

This organization of the data allows for easy querying and analysis, as well as enabling its contextualization. A summary of the data model used to create the database is presented in Figure 1.

## 3.2. Scenario Definition

The scenario definition step involves defining the assumptions and impacts of the scenario, based on the input of the end user. The assumptions represent the factors that are expected to change in the scenario, and can include production, trade, and consumption patterns. The impacts represent the factors that the user is interested in evaluating.

An assumption is defined as a set of attributes that includes an item, a metric, region, and frequency. Additionally, the assumption includes the percentage change of the metric in relation to the item in the region (and potentially a secondary region) over a specified time period, defined by the frequency. An impact is defined in a similar manner, while the attribute "change" represents the value to be predicted. A scenario can include one or multiple assumptions and one or multiple impacts. The structure of a scenario is illustrated in Figure 2.

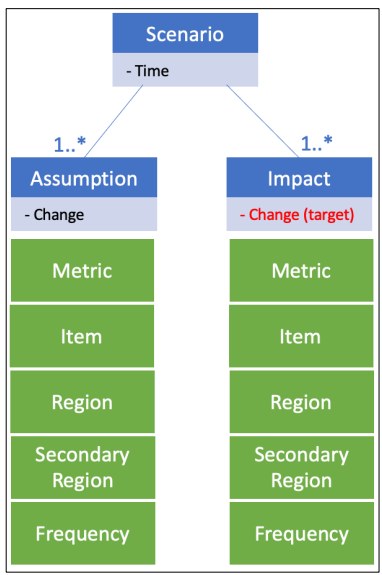

Figure 2. Scenario structure

The scenario template is a structured format that facilitates the specification of assumptions and impacts of the scenario by the end user. It comprises of two main sections: (1) Assumption specification and (2) Impact specification. The assumption specification section includes a list of assumptions, while the impact specification section includes a list of impacts. The scenario definition process is interactive and involves the interaction between the end user and the scenario definition system, which is a software tool that implements the scenario template and enables the end user to specify the assumptions and impacts of the scenario. The scenario definition system guides the end user in specifying the assumptions and impacts of the scenario by providing forms. Figures 3 presents the forms used to specify an assumption (left) and an impact (right), respectively.

Figure 3. Forms to specify an assumption (left) and an impact (right)

The system also provides feedback to the end user on the completeness and consistency of the scenario specification and enables the end user to modify the scenario specification as needed.

## 3.3. Scenario Evaluation

In this proposed approach, we aim to evaluate the potential impacts of a given scenario by utilizing time series data. The input time series data used in the approach are derived from the assumptions and impacts identified during the scenario definition phase.

The initial step of the evaluation process involves training forecasting models that can generate predictions for the impacts of the scenario. These models are trained on a subset of data from our proprietary time series food security database, which aligns with the defined assumptions and impacts. To accomplish this, we employ a Vector Autoregression (VAR) model, a widely used multivariate time series forecasting technique known for its ability to handle multiple correlated variables. This model is particularly well-suited for our application as it can handle the complex interactions between various food security indicators such as production, trade, consumption, prices, availability, access, and nutritional value. Additionally, we use a set of specific parameters such as lag, exogenous variables, and test statistics to optimize the VAR model, ensuring its ability to accurately predict future values of the series corresponding to the impacts of the scenario.

The trained VAR models employ past values of series linked to the assumptions and impacts to forecast future values of series linked to the impacts of the scenario. By doing so, they enable making predictions about the potential impacts of the scenario based on the assumptions identified during the scenario definition stage. Furthermore, we generate series values for each assumption based on the change value specified in the assumption. These generated series are then used in combination with the impact time series as input for the trained models, resulting in the final predictions of the impacts values.

To estimate the uncertainty of the predictions, we employ Monte Carlo simulation. We randomly generate different sets of input values for the VAR model, which can be different scenarios of the assumptions, and run each model multiple times, each time with a different set of input values. A normal distribution with a specific mean and standard deviation is used to generate the random input values, and a total of 5000 number of simulations are run. This allows us to generate a range of potential predictions for the impacts of the scenario on food security and to estimate the uncertainty of these predictions using measures such as the standard deviation, interquartile range and confidence intervals.

The output of the scenario evaluation step is a set of predictions for the impacts of the scenario on food security, along with an estimate of the uncertainty of the predictions, which are presented in the form of tables and figures for better interpretation. Figure 4 summarizes the evaluation workflow.

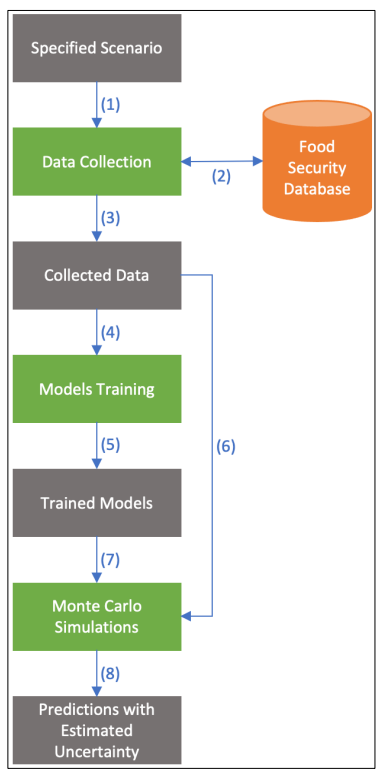

Figure 4. Summary of the evaluation workflow

## 4. RESULTS

The proposed approach for predicting the impacts of scenarios on food security was applied to multiple case studies in the United Arab Emirates (UAE). The UAE was selected as a case study due to its high dependence on food imports and the need to enhance food security in the country. In this section, we present one of these case studies, which involves the examination of a scenario of a food crisis in the UAE resulting from a disruption of the wheat supply from Russia and Ukraine, as well as an increase in food demand due to population growth.

### 4.1. Scenario Definition

The scenario definition step involved the specification of the assumptions and impacts of the scenario by the end user, using the scenario template.

The end user specified the following assumptions for the scenario: (1) a 50% reduction in the production of wheat in Russia; (2) a 100% reduction in the production of wheat in Ukraine; and (3) a 5% increase in the population of the UAE.

The end user specified the following impacts for the scenario: (1) the price of wheat in the UAE; (2) the availability of wheat in the UAE; (3) the access to wheat in the UAE; (4) the nutritional value of the diet in the UAE.

### 4.2. Analysis of the scenario evaluation results

The results of the scenario evaluation provided valuable insights into the potential impacts of scenarios on food security in the UAE. The following table shows the quantitative results of the evaluation:

Table 1: Results of the scenario execution

| Impact | Predicted change (%) | 95% CI |
|---|---|---|
| price of wheat | +15 | (12, 18) |
| the availability of wheat | +15 | (12, 16) |
| access to wheat | -15 | (-18, -12) |
| nutritional value of the diet | -10 | (-12, -8) |

The predicted increase in the price of wheat in the UAE highlights the importance of diversifying food sources and reducing dependence on imports, as well as the need for effective price stabilization mechanisms. The predicted decrease in the availability and access to wheat in the UAE highlights the need for effective food distribution systems and emergency response plans, as well as the importance of investing in local food production to increase self-sufficiency. The predicted decrease in the nutritional value of the diet in the UAE highlights the need for dietary diversification and promoting healthy eating habits.

## 5. CONCLUSION

In this paper, we proposed an approach for predicting the impacts of scenarios on food security. The approach involves two main steps: (1) scenario definition, in which the end user specifies the assumptions and impacts of the scenario using a scenario template, and (2) scenario evaluation, in which a Vector Autoregression (VAR) model is used in combination with Monte Carlo simulation to generate predictions for the impacts of the scenario based on the defined assumptions and impacts. We applied the proposed approach to a case study of food security in the United Arab Emirates (UAE), and found that the approach was able to predict the potential impacts of a scenario involving a disruption in the wheat supply from Russia and Ukraine and an increase in population in the UAE. The results of the case study provide valuable insights into the potential impacts of scenarios on food security and can inform decision-making processes related to enhancing food security in the UAE and other countries.

The proposed approach can be used to predict the potential impacts of other scenarios on food security, such as natural disasters, climate change, or economic downturns. The approach can be used to identify potential vulnerabilities and areas for improvement, and to develop effective strategies for enhancing food security. However, it is important to note that the approach does have some limitations, such as the availability of accurate and reliable data and the ability of the end user to specify realistic and plausible assumptions and impacts for the scenario.

Overall, the proposed approach provides a valuable tool for predicting the potential impacts of scenarios on food security and can be used to inform decision-making processes related to enhancing food security. However, it is important to consider the limitations of the approach and to conduct further research in order to validate the findings and to explore the potential applications of the approach in other contexts and scenarios.